\documentclass{article}

\usepackage[preprint, nonatbib]{neurips_2025}

\usepackage{wrapfig}
\usepackage[utf8]{inputenc} % allow utf-8 input
\usepackage[T1]{fontenc}    % use 8-bit T1 fonts
\usepackage{hyperref}       % hyperlinks
\usepackage{url}            % simple URL typesetting
\usepackage{booktabs}       % professional-quality tables
\usepackage{amsfonts}       % blackboard math symbols
\usepackage{nicefrac}       % compact symbols for 1/2, etc.
\usepackage{microtype}      % microtypography
\usepackage{multirow}
\usepackage{makecell}
\usepackage[numbers,sort&compress]{natbib}
\usepackage{graphicx}
\usepackage{subcaption}
\usepackage[table]{xcolor} 

\usepackage[most]{tcolorbox}
\usepackage{placeins}

\newtcolorbox{promptbox}{
  colback=gray!10,      % light gray background
  colframe=gray!50,     % darker gray border
  boxrule=0.5pt,        % border thickness
  arc=2pt,              % slightly rounded corners
  left=3pt,             % inner left padding
  right=3pt,            % inner right padding
  top=3pt,
  bottom=3pt,
  breakable             % allow it to split across pages if needed
}

\usepackage{graphicx}
\usepackage[inline]{enumitem}
\usepackage{amsmath}
\usepackage{amssymb}

\title{\smash{\raisebox{-1.7em}{\includegraphics[width=2.9em,height=2.9em]{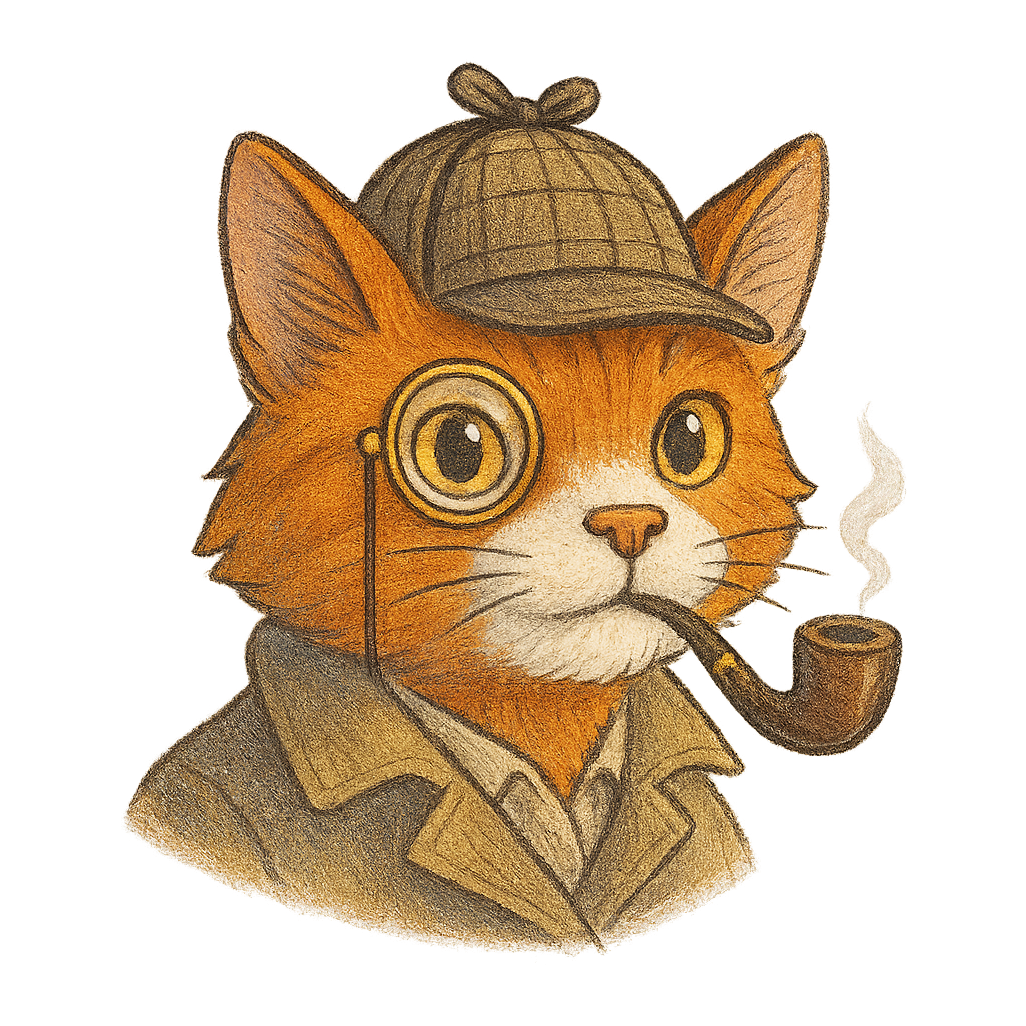}}}DAVID-\raisebox{-0.1em}{\includegraphics[width=0.9em,height=0.9em]{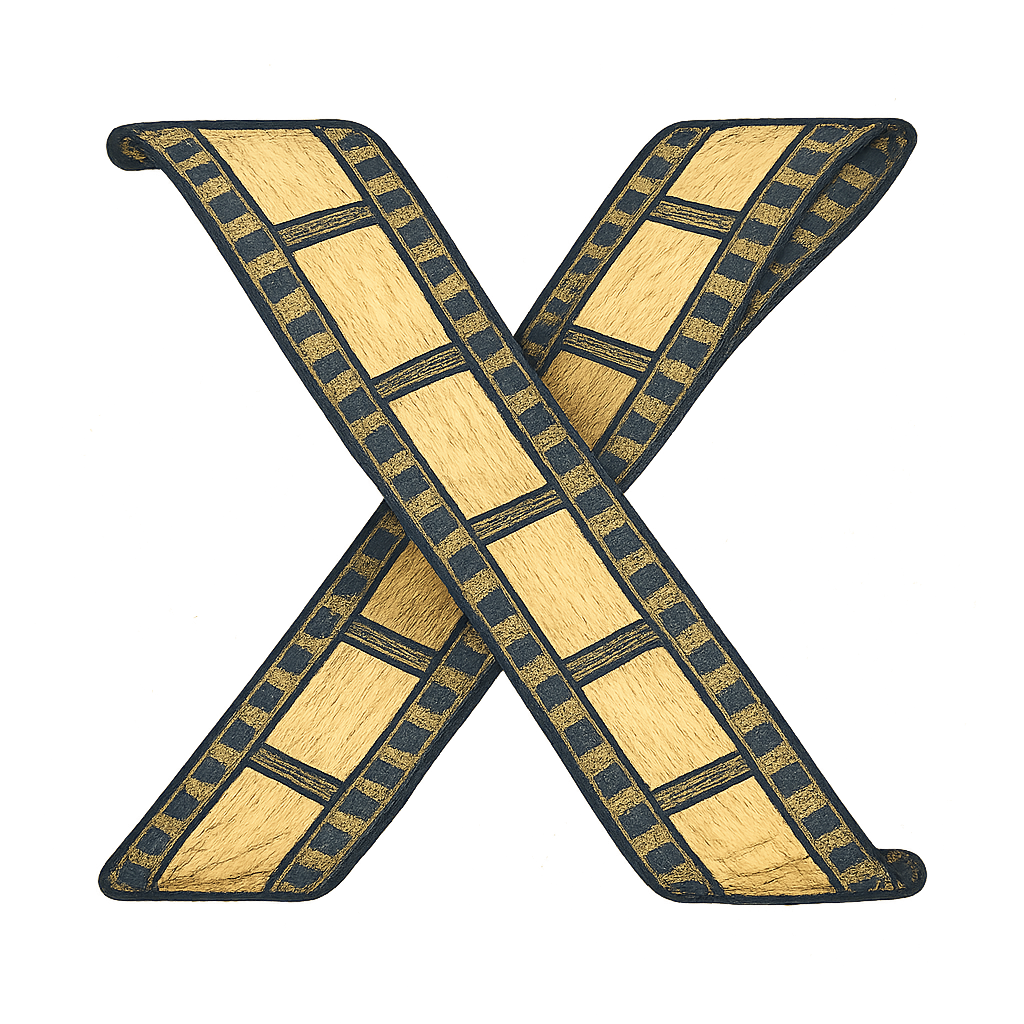}}R1: Detecting AI-Generated Videos with Explainable Reasoning}
\author{%
Yifeng~Gao$^{1}$\quad
Yifan~Ding$^{1}$\quad
Hongyu~Su$^{1}$\quad
Juncheng~Li$^{1}$\quad
Yunhan~Zhao$^{1}$\quad
Lin~Luo$^{1}$\\[2pt]
\bfseries
Zixing~Chen$^{1}$\quad
Li~Wang$^{1}$\quad
Xin~Wang$^{1}$\quad
Yixu~Wang$^{1}$\quad
Xingjun~Ma$^{1}$\thanks{Corresponding author}\quad
Yu\mbox{-}Gang~Jiang$^{1}$\\[6pt]
$^{1}$Fudan University
}

\begin{document}

\maketitle

\begin{abstract}
As AI-generated video becomes increasingly pervasive across media platforms, the ability to reliably distinguish synthetic content from authentic footage has become both urgent and essential. Existing approaches have primarily treated this challenge as a binary classification task, offering limited insight into where or why a model identifies a video as AI-generated. However, the core challenge extends beyond simply detecting subtle artifacts; it requires providing fine-grained, persuasive evidence that can convince auditors and end-users alike.
To address this critical gap, we introduce \textit{DAVID-X}, the first dataset to pair AI-generated videos with detailed defect-level, temporal–spatial annotations and written rationales. Leveraging these rich annotations, we present \textbf{DAVID-XR1}, a video–language model designed to deliver an interpretable chain of visual reasoning—including defect categorization, temporal–spatial localization, and natural language explanations. This approach fundamentally transforms AI-generated video detection from an opaque black-box decision into a transparent and verifiable diagnostic process.
We demonstrate that a general-purpose backbone, fine-tuned on our compact dataset and enhanced with chain-of-thought distillation, achieves strong generalization across a variety of generators and generation modes. Our results highlight the promise of explainable detection methods for trustworthy identification of AI-generated video content.

\end{abstract}

\section{Introduction}

With the rapid advancement of video generation models \cite{blattmann2023stable, blattmann2023align, he2022latent, hong2022cogvideo, wang2023modelscope, chen2023videocrafter1, zheng2024open, lin2024open}, AI-generated videos has evolved from a tentative experiment into a mainstream creative medium. Today, a single line of text or a reference image can be transformed into a complete video, dramatically lowering the barrier to content creation and inundating social media platforms with AI-generated video clips. State-of-the-art (SOTA) models such as Veo 2 \cite{veo2}, Sora \cite{videoworldsimulators2024}, Kling \cite{kling_text2video}, and Hunyuan \cite{kong2024hunyuanvideo} now produce videos with unprecedented visual fidelity and temporal coherence, making their outputs nearly indistinguishable from both naturally recorded footage and handcrafted animation. This deceptive realism significantly increases the risk of misuse and overwhelms current content moderation systems, underscoring the urgent need for specialized detection methods for AI-generated videos.

Recent research has approached AI-generated video detection either as a binary ``real vs. AI-generated'' classification task or as a model-attribution task \cite{ma2025detectingaigeneratedvideoframe, chen2024demambaaigeneratedvideodetection, bai2024ai, kundu2024towards, he2024exposing, liu2024turns, vahdati2024beyond}, achieving impressive accuracy and establishing a foundation for authenticity verification. However, these classifiers cannot provide explanations for why a video is classified as AI-generated, making their decisions difficult to interpret and trust.
Social media and online video platforms increasingly require that every clip flagged as ``suspected AI-generated content''  be supported by concrete, explainable evidence, enabling human moderators to identify and clearly communicate the basis for such flags. The central challenge, therefore, is to provide detection results that are supported by precise spatio-temporal localization and clear, human-readable explanations.

To address this need, we first construct and release \textit{DAVID-X}, a small-scale yet richly annotated dataset for AI-generated video detection. It consists of 416 AI-generated clips from 15 mainstream generators and 337 real videos with matched semantic content. For each synthetic clip, the annotation process mirrors a human reviewer’s reasoning: the clip is first matched to the most similar category of human-made footage, and each generation flaw is meticulously labeled by defect type \cite{liu2023fetv, huang2024vbench}, frame range, 2D coordinates, and accompanied by a concise natural language explanation. These hand-marked coordinates can be directly used with SAM 2 \cite{ravi2024sam} to generate pixel-level masks of flawed regions. Additionally, we provide explanatory annotations for real videos, highlighting the cues supporting their classification as authentic. By combining a curated set of trending prompts, diverse video genres, and richly interpretable labels, \textit{DAVID-X} offers a dedicated training corpus for video-language multimodal models (VLMMs) \cite{zhang2024llava, Qwen2.5-VL, chen2024internvl, wang2024internvideo2, yao2024minicpm, li2024llava, team2023gemini}, aimed at both accurate and explainable AI-generated video detection.

% \begin{figure}[!ht]
%   \centering
%   \includegraphics[width=0.9\linewidth]{Fig/Illustration.pdf}
%   \caption{Illustration of David X1.}
%   \label{Fig:Illustration}
% \end{figure}

\begin{wrapfigure}{r}{0.5\textwidth}
  \vspace{-1.1em}          
  \centering
  \includegraphics[width=\linewidth]{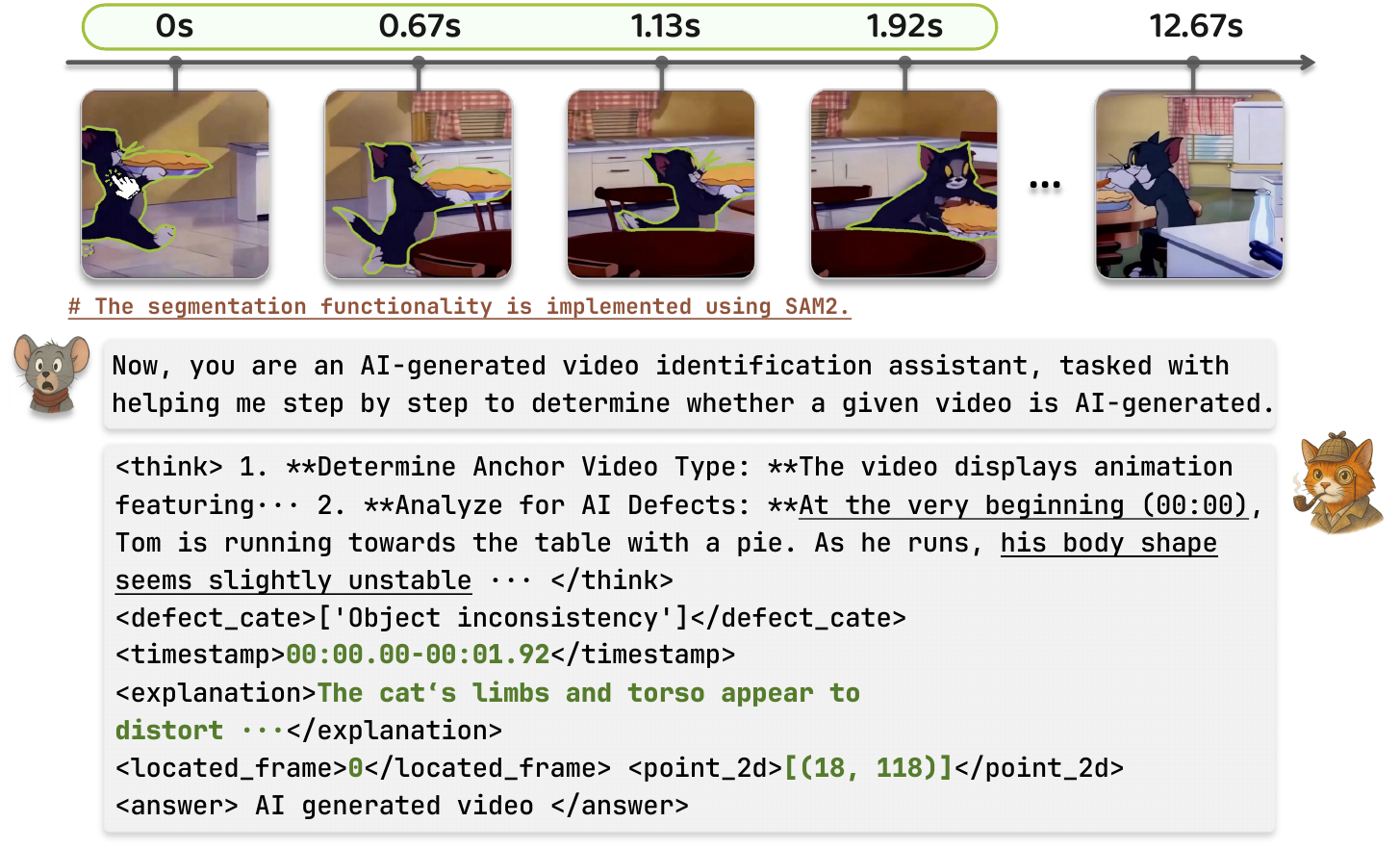}
  \caption{An illustration of \textbf{DAVID-XR1}’s explainable reasoning capability.}
  \label{fig:illustration}
  % \vspace{-1.0em}           
\end{wrapfigure}

Inspired by the visual reasoning principle of “extract evidence first, then reason step-by-step” \cite{han2024videoespresso, shao2024visual, wu2024v, wang2024videocot}, we begin by distilling Chain-of-Thought (CoT) traces from the proprietary Gemini 2.5 Pro, restructuring video forensics into five explicit stages: \emph{defect discovery}, \emph{spatial–temporal localization}, \emph{defect categorization}, \emph{defect description}, and \emph{final verdict}. Building on these traces, we perform \textbf{CoT-aware SFT} on an open-source vision–language backbone. Specifically, we attach a lightweight binary classifier to the final CoT hidden state to predict “real” vs. “AI-generated,” training it alongside the model’s standard language-generation objective. This auxiliary supervision tightly integrates explanation with classification, and our experiments demonstrate that such joint training significantly improves interpretability.
The resulting model, \textbf{DAVID-XR1} (illustrated in Figure~\ref{fig:illustration}), makes its entire decision chain transparent, boosts backbone accuracy on out-of-domain (OOD) AI-generated video detection from 26.7\% to 76.7\%, and achieves 54.7\% explanation precision—bringing its reasoning and explanation capabilities on par with Gemini 2.5 Pro and GPT-4.1.

% we perform \textbf{CoT-aware SFT} on an open-source vision–language backbone: during fine-tuning, a lightweight binary classifier is appended to read the model’s final hidden state of CoT process and predict whether the video is real or AI-generated. Its cross-entropy loss is optimized jointly with the usual language-generation loss over the entire output sequence. 

In summary, our main contributions are as follows:
\begin{itemize}
\item We release \textit{DAVID-X}, the first dataset to mirror the human decision-making workflow by providing fine-grained, spatio-temporal defect annotations for AI-generated videos, along with parallel explanatory labels for semantically matched real videos.

\item Building on \textit{DAVID-X} and visual CoT distillation, we introduce \textbf{DAVID-XR1}, a vision-language multimodal model that not only detects AI-generated videos but also generates an explicit, interpretable reasoning chain.

\item We empirically demonstrate that \textbf{DAVID-XR1} generalizes well to out-of-domain AI-generated videos and validate the effectiveness of our \emph{CoT-aware SFT} strategy, which jointly optimizes reasoning and classification, resulting in significant improvements in both detection accuracy and explanation quality.
\end{itemize}

\section{Related Work}

\paragraph{AI-Generated Video Detection}
Recent research in AI-generated video detection has primarily focused on distinguishing synthetic videos from authentic ones. Early approaches leveraged CNN architectures for feature extraction and classification \cite{vahdati2024beyond, bai2024ai}. For instance, \cite{vahdati2024beyond} employed CNNs to extract synthetic traces for video-level discrimination, while \cite{bai2024ai} incorporated optical flow maps as an auxiliary input to enhance classification accuracy. Transformer-based methods have also gained traction: DeCoF \cite{ma2025detectingaigeneratedvideoframe} uses a pre-trained CLIP model to emphasize inter-frame inconsistencies, and UNITE \cite{kundu2024towards} introduces an Attention-Diversity (AD) Loss to encourage spatial diversity among attention heads. Hybrid architectures, such as \cite{he2024exposing}, combine CNNs and Transformers to capture both local motion features and global appearance variations. DIVID \cite{liu2024turns} extracts RGB features and Diffusion Reconstruction Error (DIRE) with a CNN, feeding these into an LSTM for temporal modeling and classification. Large-scale efforts like DeMamba \cite{chen2024demambaaigeneratedvideodetection} further advance the field by training detail-aware modules on million-scale datasets.

Distinct from prior work, which treats AI-generated video detection as a classification problem, our research emphasizes explainable detection by adopting a VLMM approach that aims not only for high accuracy but also for detailed explanations. 
By enhancing the model’s ability to recognize and articulate specific artifacts in generated videos, our approach seeks to provide concrete, discriminative evidence and fulfill the growing demand for explainable results in AI-generated video detection.

\paragraph{Visual CoT}
With the rapid progress of multimodal models, an increasing number of studies are extending CoT reasoning from text-only settings to enhance multimodal and visual understanding. Some approaches decompose the reasoning process into sequential stages: Multimodal-CoT \cite{zhang2023multimodal} generates rationales from visual inputs, while CCoT \cite{mitra2024compositional} constructs scene graphs to guide subsequent inference. Others exploit vision’s spatio-temporal grounding, such as $V^*$\cite{wu2024v}, which iteratively searches for and localizes relevant cues to improve reasoning accuracy. Visual CoT \cite{shao2024visual} introduces datasets with annotated critical regions, enabling models to focus on the most salient visual elements, and VideoEspresso \cite{han2024videoespresso} incorporates spatio-temporal grounding to further boost video understanding. Meanwhile, VoT \cite{fei2024video} emulates a human-like progression from low-level pixel perception to high-level semantic reasoning.
Inspired by these advances, our approach decomposes AI-generated video detection into a stepwise, human-interpretable reasoning process. We start by localizing generated defects, categorize and describe them, and ultimately reach a verdict on whether the video is AI-generated, grounding our decision in clearly identified evidence.

\section{Explainable AI-Generated Video Detection}
In this section, we present our approach for explainable detection of AI-generated videos. We begin by introducing \textit{DAVID-X}, a finely annotated dataset tailored for this task. Next, we describe the training strategy for \textbf{DAVID-XR1}, an explicitly interpretable VLMM model. Finally, we conclude with key insights gained from this training paradigm.

\begin{figure}[!t]
  \centering
  \includegraphics[width=1.0\linewidth]{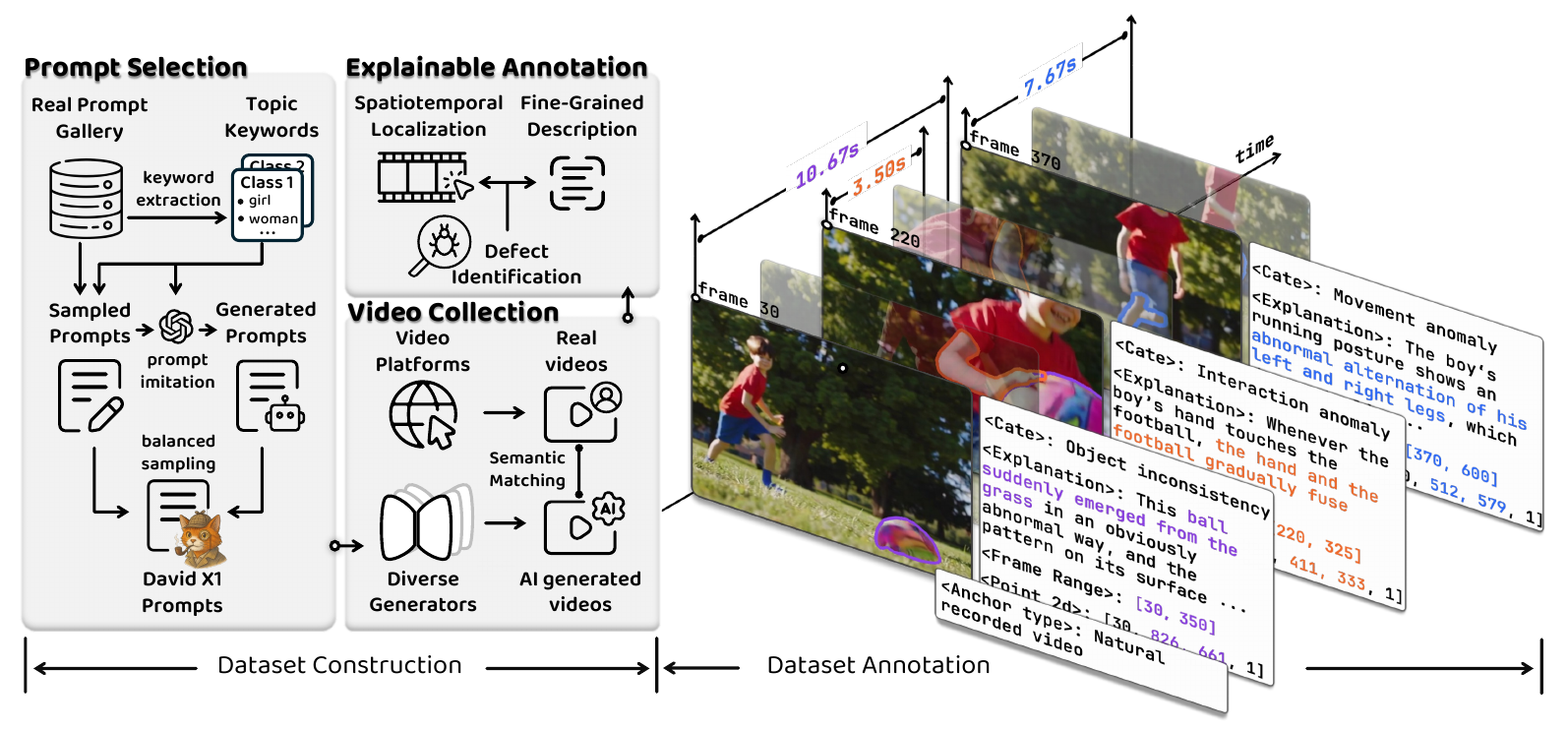}
  \caption{Overview of the data construction and annotation workflow for our \textit{DAVID-X} dataset.}
  \vspace{-1.0em}
  \label{Fig:Illustration}
\end{figure}

\subsection{Dataset Construction}
To equip VLMMs with explainable reasoning capabilities for AI-generated video detection, we first construct a dedicated dataset. We introduce \emph{DAVID-X}, a corpus of 747 AI-generated and real videos, each paired with annotations that closely follow the human decision-making process. The creation of \emph{DAVID-X} involves a three-stage pipeline: 1) \emph{prompt selection}, 2) \emph{video generation and collection}, and 3) \emph{human-style detection annotation}.

\subsubsection{Prompt Selection}
One essential requirement for the dataset is that it should represent real-world video generation scenarios using a diverse set of authentic prompts. 
To achieve this, we select prompts from VidProM~\cite{wang2024vidprom}—a large gallery of real user queries that already includes text-embedding-3-large vectors—to generate the videos.
We use UMAP~\cite{mcinnes2018umap} to project these embeddings into three dimensions, then apply K-Means~\cite{burkardt2009k} clustering with $k = 80$ to the low-dimensional points. The 30 largest clusters together cover over 89\% of the prompt pool while maintaining clear separation, so we focus our sampling on these clusters.

Within each cluster, we apply TF-IDF to identify 10 keywords that are frequent within the cluster but rare elsewhere, thereby capturing the thematic essence and maximizing inter-cluster distinctiveness. Using these keywords, we select 30 representative prompts per cluster—each containing at least two cluster-specific keywords—to build our sampled prompt set. To further enhance realism and reflect the common practice of generators rewriting user queries with LLMs, we generate additional prompts: for each cluster, GPT-o1~\cite{jaech2024openai} is tasked with crafting 30 new prompts based on the keyword list and three sampled prompts as exemplars. This two-tiered approach produces a balanced, semantically diverse prompt collection that mirrors real-world generation scenarios while keeping the dataset size manageable.

Additionally, drawing on established methods in video dataset creation, we employ GPT-4o~\cite{hurst2024gpt} to tag each prompt with one or more of eight major spatial-content categories: \emph{people}, \emph{animals}, \emph{vehicles}, \emph{plants}, \emph{artifacts}, \emph{food}, \emph{buildings}, and \emph{scenery}. We then perform Monte Carlo sampling to select 100 prompts, jointly balancing (i) the 30 popular topic clusters, (ii) the spatial-content labels, and (iii) the sampled versus generated split. This procedure yields a concise yet highly diverse prompt set, which serves as the prompt collection for \textit{DAVID-X}, ensuring broad coverage of topics, visual entities, and real-world usage patterns.

\subsubsection{Video Generation and Collection}
With the selected prompt list from the previous step, we next construct the video corpus for \emph{DAVID-X}. To capture the broad stylistic and quality spectrum of current text-to-video models, we generated clips using 9 commercial models, including Sora~\cite{Sora}, Pika v1.5~\cite{Pika}, Gen-3 Alpha~\cite{Gen-3}, KLING 1.5~\cite{KLING}, Jimeng 1.2, Jimeng 2.0~\cite{Jimeng}, PixVerse V3~\cite{PixVerse}, Vidu 1.5~\cite{Vidu}, and Dream Machine~\cite{Bream_Machine}, and 6 open-source models ranging from early to the most recent releases: HunyuanVideo~\cite{kong2024hunyuanvideo}, Open-Sora Plan v1.3~\cite{lin2024open}, Open-Sora 1.2~\cite{zheng2024open}, CogVideoX 1.5-5B~\cite{yang2024cogvideox}, VideoCrafter 2~\cite{chen2024videocrafter2}, and Modelscope~\cite{wang2023modelscope}. 
From these sources, we curated 416 AI-generated video clips, covering a wide variety of formats: low frame rates (10 fps) to smooth (30 fps), brief micro-shots (less than 2 seconds) to extended 20-second scenes, and resolutions from coarse $256 \times 256$ to crisp full HD $1920 \times 1080$. This diversity ensures that the dataset captures the full range of artifacts and stylistic nuances present across state-of-the-art generation engines.

\begin{figure}[htbp]
  \centering
  \begin{minipage}[t]{0.54\textwidth}
    \vspace{0pt}
    \centering
    \captionof{table}{Counts of videos generated by 15 models and real videos.}
    \label{tab:video_source_counts}
    \resizebox{\linewidth}{!}{%
      \begin{tabular}{l r l r}
        \toprule
        Video Source           & Count & Video Source       & Count \\
      \midrule
      Open-Sora Plan v1.3~\cite{lin2024open}    & 24    & Sora~\cite{Sora}               & 30    \\
      Modelscope~\cite{wang2023modelscope}             & 31    & KLING 1.5~\cite{KLING}          & 33    \\
      Open-Sora 1.2~\cite{zheng2024open}          & 34    & Vidu 1.5~\cite{Vidu}           & 29    \\
      Dream Machine~\cite{Bream_Machine}          & 26    & HunyuanVideo~\cite{kong2024hunyuanvideo}       & 32    \\
      Gen-3 Alpha~\cite{Gen-3}            & 36    & Jimeng 1.2~\cite{Jimeng}         & 20    \\
      Jimeng 2.0~\cite{Jimeng}             & 27    & PixVerse V3~\cite{PixVerse}        & 32    \\
      CogVideoX1.5-5B~\cite{yang2024cogvideox}        & 16    & Pika 1.5~\cite{Pika}           & 33    \\
      VideoCrafter2~\cite{chen2024videocrafter2}          & 13    & Real~\cite{SafeVid-350K}               & 337   \\
        \bottomrule
      \end{tabular}%
    }
  \end{minipage}
  \hfill
  \begin{minipage}[t]{0.44\textwidth}
    \vspace{0pt}
    \centering
    \includegraphics[width=\linewidth]{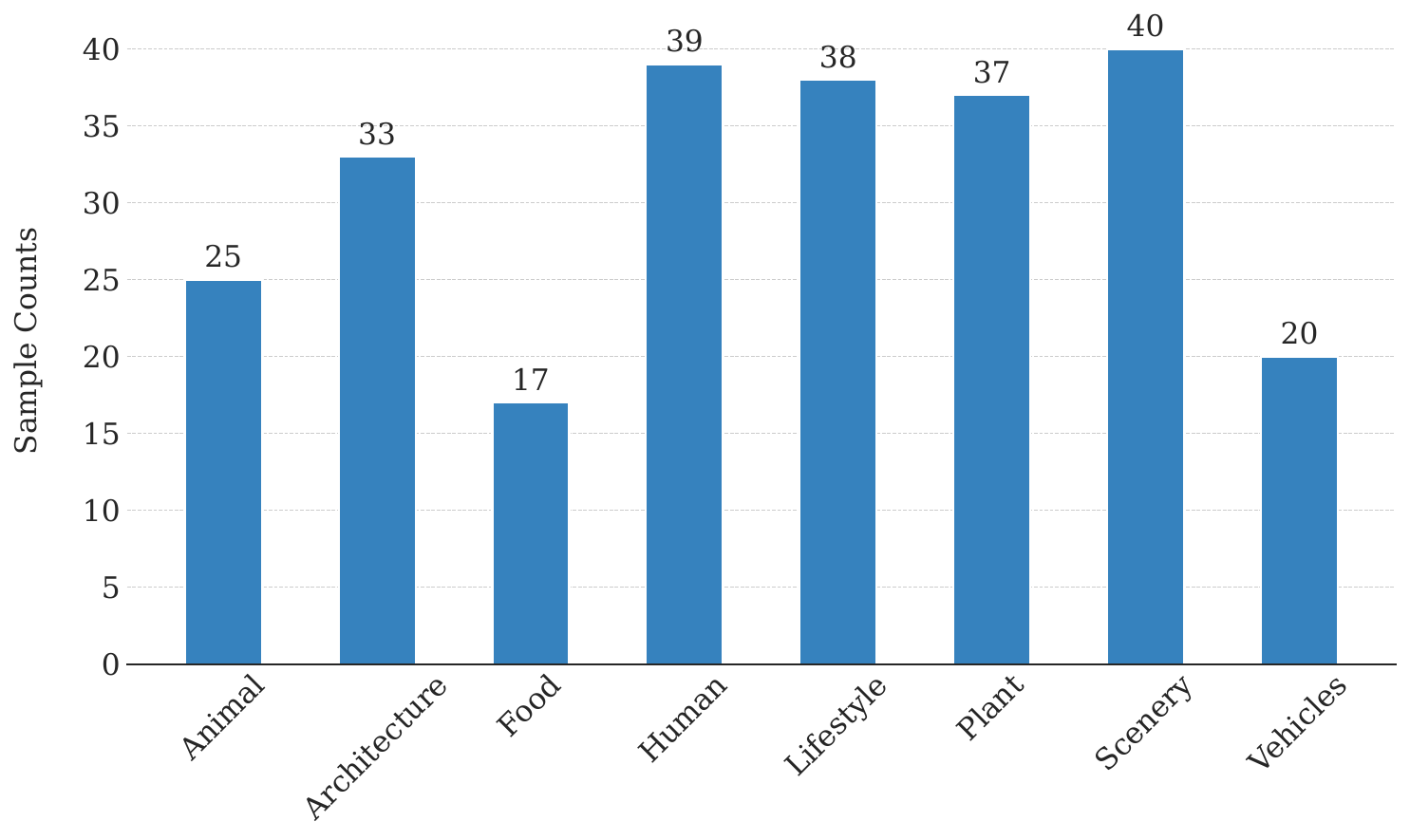}
    \captionof{figure}{Statistics of DAVID-X dataset.}
    \label{plot:video_cate}
  \end{minipage}
\end{figure}

For real videos, we first manually-curated 147 clips from major video platforms using the same prompts as for the generated videos. To further increase diversity, we include 190 segments from the scene-diverse subset of InternVid~\cite{SafeVid-350K}. These videos were non-overlappingly cropped into 5s, 10s, 20s, and 30s chunks, and only those with a VideoCLIP~\cite{xu2021videoclip} embedding cosine similarity of $\geq 0.22$ to any prompt were retained. This process yielded a total of 331 real videos. By carefully matching prompts and applying semantic filtering, we minimize topical bias, encouraging models to focus on genuine generation defects when constructing their reasoning chains.

\subsubsection{Human-Style Detection Annotation}
After assembling the video corpus, we developed a dedicated annotation tool that closely mirrors the workflow of human inspectors reviewing AI-generated footage. Human raters assess a clip’s authenticity by identifying visible generation flaws—a fundamentally different approach from methods that rely solely on visual models extracting high-level features for classification. Explainable detection of AI-generated videos requires more than a simple binary verdict; the model must also provide concrete evidence that auditors and end-users can visually verify. Accordingly, our dataset records not only the authenticity label but also the specific, human-recognizable defects that distinguish synthetic clips from real ones, along with clear explanations of how these defects support the final decision.

\paragraph{Anchor-Type Assignment}
To accurately identify AI-generated content, we define AI-generated videos as clips produced by a generative model, while real videos fall into two categories:
\begin{enumerate*}[label=(\roman*)]
\item \emph{Naturally recorded videos}: footage captured directly by a camera with minimal post-processing.
\item \emph{Handcrafted videos}: content involving substantial post-production, such as film VFX, animation, or game footage.
\end{enumerate*}
During annotation, each synthetic clip is first anchored to the real-video category it most closely resembles. This anchoring is essential, as the tolerance for artifacts varies: handcrafted footage is expected to diverge more from physical realism, and thus requires stronger evidence before being labeled as AI-generated. The selected anchor is stored in an \texttt{<anchor type>} tag, ensuring that subsequent defect labels are evaluated against the appropriate baseline. This strategy improves annotation consistency and helps reduce false positives.

\begin{table}[htbp]
  \centering
  \caption{Counts of detects for video anomalies}
  \label{tab:defect_counts}
  \resizebox{\textwidth}{!}{%
    \begin{tabular}{l|cccccc}
      \toprule
      Defect & Object Inconsistency & Texture Jitter & Interaction Anomaly & Movement Anomaly & Space Anomaly & Lighting Anomaly \\
      \midrule
      Count  & 888                  & 300            & 269                  & 71               & 22            & 12                 \\
      \bottomrule
    \end{tabular}%
  }
\end{table}

\paragraph{Defect Categorization}
Once a defect is identified in an AI-generated clip, we assign it to one or more defect classes. Guided by the taxonomies established in recent AI-generated video benchmarks, we adopt 6 explainability-oriented categories:

\begin{itemize}
\item \textbf{Object Inconsistency}: Failure of an object to maintain consistent attributes, such as shape, size, color, or fine details, across frames, resulting in illogical morphing, distortion, or loss of detail.
\item \textbf{Texture Jitter}: High-frequency flickering or drifting of surface textures over time, manifested as grid-like noise, “heat-haze” ripples, or crawling patterns—distinct from compression artifacts or uniform blur.
\item \textbf{Interaction Anomaly}: Unnatural blending, sticking, or partial fusion when visually similar objects overlap or touch, leading to blurred boundaries.
\item \textbf{Movement Anomaly}: Motions or trajectories that appear jerky, distorted, or physically/biologically implausible.
\item \textbf{Space Anomaly}: Spatial inconsistencies revealed by camera motion, such as mismatched parallax between foreground and background, or seams where newly exposed regions fail to align.
\item \textbf{Lighting Anomaly}: Shadows or highlights that contradict the scene’s light sources, or implausible flickering and intensity shifts over time.
\end{itemize}

For clips anchored to handcrafted video, we apply a looser standard for Movement, Space, and Lighting anomalies, as such footage is not expected to adhere strictly to real-world physics. Each detected flaw is recorded in the \texttt{<defect cate>} tag, enabling fine-grained supervision that helps models learn the rationale behind each video’s flag, thereby enhancing both detection accuracy and explanation quality.

\paragraph{Spatio-Temporal Localization}
For every detected defect, we annotate both \textbf{when} and \textbf{where} it occurs. The affected frames are recorded in the \texttt{<frame range>} tag and converted to human-readable spans in \texttt{<timestamp>} using the video’s frame rate, a format compatible with the temporal grounding capabilities of current VLMMs. This temporal annotation is crucial for capturing the often fleeting nature of AI-generated defects.
To precisely localize the defect region, we leverage SAM2’s point-to-mask interface. For each defect, we annotate a sequence of 2D prompts, recorded as quadruples $\texttt{[frame, x, y, label]}$ in the \texttt{<point>} tag, where \textit{frame} is the annotated frame index, $(x, y)$ are the pixel coordinates, and \textit{label} indicates a positive or negative prompt for SAM2. While these points enable precise mask generation (especially useful if a VLMM agent interacts with SAM2), we retain only the positive point on the first affected frame during training, aligning with the coarse localization capabilities of most general-purpose VLMMs. By combining concise temporal spans with sparse, actionable point prompts, our annotations offer lightweight yet effective supervision for both temporal and spatial reasoning.

\paragraph{Explanation}
Each defect is also accompanied by a textual explanation, stored in the \texttt{<explanation>} tag, which articulates why the flaw indicates AI generation. This completes the annotation for a single defect. Each video may include multiple such entries, providing the model with broader evidence and fostering robust reasoning, even on out-of-domain content.
By integrating anchor-type selection, defect categorization, precise spatio-temporal localization, and human-interpretable explanations into a unified workflow, \textit{DAVID-X} establishes the first fine-grained dataset designed to train models not only to detect AI-generated videos but also to clearly articulate the visual evidence underlying their decisions.

\subsection{Model Training}

Inspired by recent studies on video CoT reasoning, which demonstrate that incorporating spatiotemporal localization of key objects into the reasoning chain significantly enhances video understanding, we integrate our fine-grained annotations within a visual CoT framework. In this approach, explainable AI-generated video detection is formulated as a three-step process:
\begin{enumerate*}[label=(\roman*)]
\item \emph{searching for generation defects},
\item \emph{localizing them in space and time}, and
\item \emph{determining whether each defect constitutes decisive evidence}.
\end{enumerate*}
The model scans the entire clip for such cues, aggregates the findings, and delivers its verdict—an end-to-end workflow we refer to as \textit{explainable reasoning}.

\paragraph{Visual-CoT Distillation}
To obtain high-quality visual CoT, we distill them from Gemini 2.5 Pro, a model renowned for its advanced vision and reasoning capabilities. Each video is presented to Gemini along with \textit{DAVID-X}’s ground-truth evidence (anchor type, defect class, timestamp, point, and explanation) as well as formal definitions for each sub-label. The prompt instructs Gemini to proceed step-by-step within \texttt{<think>}, explicitly deriving (rather than inventing) the ground-truth cues. All evidence tags must match the ground truth exactly, except for \texttt{<explanation>}, which may be expanded upon by the model. These distilled traces transform \textit{DAVID-X}’s discrete annotation cues into complete visual CoTs, which are then used to fine-tune a VLMM, imparting robust and interpretable reasoning abilities for AI-generated video detection.

\begin{wrapfigure}{r}{0.4\textwidth}
  \vspace{-2.0em}          
  \centering
  \includegraphics[width=\linewidth]{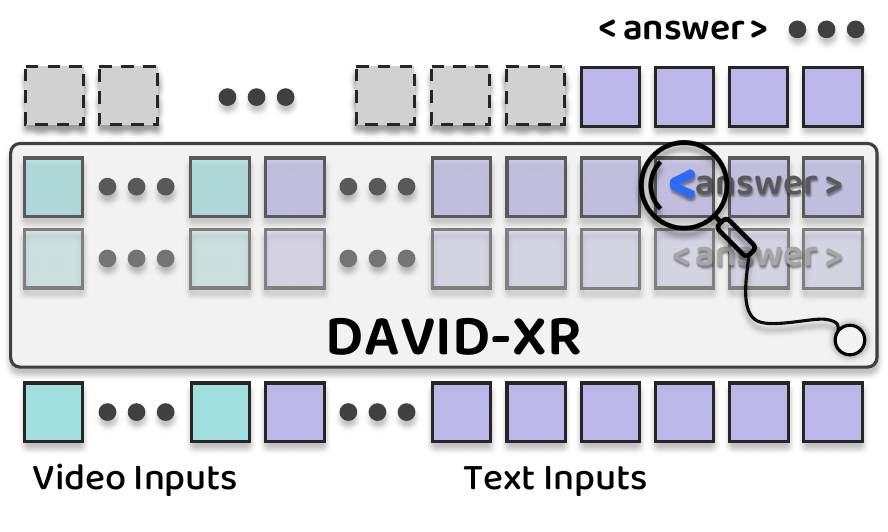}
  \caption{Illustration of CoT-aware SFT.}
  \label{fig:cot_loss}
  \vspace{-2.0em}           
\end{wrapfigure}
\paragraph{CoT-Aware SFT}
We apply supervised fine-tuning (SFT) to adapt a pretrained VLMM to our task: given a video, the model generates a coherent reasoning chain within \texttt{<think></think>}, summarizes the evidence in \texttt{<evidence></evidence>}, and concludes with a binary verdict in \texttt{<answer></answer>}. While optimizing the standard language model loss on this formatted sequence equips the model with the appropriate narrative style and tag structure, it does not yet directly supervise the reasoning process itself.

To align the model’s internal reasoning with the ground-truth decision, we introduce a lightweight binary classifier. During inference, the model’s output sequence concludes with \texttt{<answer>···<answer>}. During training, we extract the final hidden state immediately preceding this token, input it to the classifier, and predict whether the clip is AI-generated $(1)$ or real $(0)$. This additional classification head ensures that the hidden representation generated by the CoT reasoning process is closely tied to the final decision, encouraging the model to internalize and utilize the underlying evidence rather than simply replicating it. Consequently, we jointly optimize the language modeling objective $\mathcal{L}_\mathrm{LM}$ and the classification objective $\mathcal{L}_\mathrm{CLS}$:

\begin{equation}
\mathcal{L} = 
\alpha\,\underbrace{\Bigl(-\sum_{t=1}^T \log p_\theta(y_t \mid y_{<t}, \mathbf{x})\Bigr)}_{\mathcal{L}_{\mathrm{LM}}}
\;+\;
\beta\,\underbrace{\Bigl(-y\log\hat p_\theta - (1-y)\log(1-\hat p_\theta)\Bigr)}_{\mathcal{L}_{\mathrm{CLS}}}\mathstrut.
\end{equation}
% \begin{equation}
% \hat p_\theta = \sigma\bigl(\mathbf{w}^\top \mathbf{h}_{k} + b\bigr),
% \quad
% k = \text{pos}(\langle\!\text{answer}\!\rangle) - 1.
% \end{equation}

Here, the total loss is defined as $\mathcal{L} = \alpha \mathcal{L}_{\mathrm{LM}} + \beta \mathcal{L}_{\mathrm{CLS}}$, where $\mathcal{L}_{\mathrm{LM}}$ is the negative log-likelihood of the full \texttt{<think>…<evidence>…<answer>} sequence under model parameters $\theta$. The classification loss, $\mathcal{L}_\mathrm{CLS}$, is the binary cross-entropy on the sigmoid output $\hat p_\theta = \sigma\bigl(\mathbf{w}^\top \mathbf{h}{k} + b\bigr)$, where $\mathbf{h}{k}$ is the hidden state immediately preceding the \texttt{<answer>} token and $y \in {0, 1}$ is the ground-truth label for real versus AI-generated video. The scalars $\alpha$ and $\beta$ control the balance between the language modeling and classification objectives.

Jointly minimizing $\mathcal{L}$ teaches the model both \textbf{how to explain} (through the language loss) and \textbf{how to decide} (through the classification loss) in a single pass. As shown experimentally in Section~\ref{sec:ablation}, this CoT-aware SFT approach yields the best improvements in explanation capability compared to language-only fine-tuning. This training approach produces \textbf{DAVID-XR1}, an explainable-reasoning VLMM capable of interpretable AI-generated video detection.

\paragraph{Other Explorations}

We also explored the addition of a reinforcement learning (RL) phase—using either Direct Preference Optimization (DPO) or Group Relative Policy Optimization (GRPO)—after SFT to fruther improve \textbf{DAVID-XR1}’s reasoning capabilities. However, our task requires a fixed output format (\texttt{<think>…<evidence>…<answer>}, including defect categories, anchors, timestamps, and spatial points), and the SFT-tuned model seldom generates new reasoning paths during RL. Since RL typically improves models by optimizing their sampling distributions \cite{yue2025does}, this fixed reasoning structure limits the potential benefits. As a result, we restrict training to SFT and leave the exploration of CoT diversity and RL-based improvements for future work.

\section{Experiment}

\subsection{Experimental Setup}

\paragraph{Implementation Details}
Our \textbf{DAVID-XR1} model was trained on eight NVIDIA A800-SXM4-80GB GPUs, using SOTA general-purpose VLMM Qwen2.5-VL-7B~\cite{Qwen2.5-VL} as the base model for fine-tuning. We adopted LLaMA-Factory \cite{zheng2024llamafactory} as the foundational training framework for our experiments. Full fine-tuning was performed on both the Vision Tower and LLM components, using a learning rate of $1.0 \times 10^{-5}$ and a cosine learning rate scheduler.

% In our CoT-aware SFT, the weight parameters for the language and classification objectives, $\alpha$ and $\beta$, were both set to 0.5.

\paragraph{Evaluation Benchmark}
To simulate realistic detection scenarios, we constructed a 90-video test set encompassing multiple generalization challenges. This set includes two next-generation commercial engines—Kling 2.0~\cite{KLING} and Pika v2.2~\cite{Pika} (successors to Kling 1.5 and Pika v1.5 used in training)—the fully out-of-domain Veo2 model~\cite{veo2}, and the autoregressive MAGI-1~\cite{magi1}. For each of these four text-to-video systems, we crafted 15 prompts: three are manually designed (anchored to “natural recorded video,” “handcrafted video,” and multi-shot scenes), and GPT-o3~\cite{GPT-o3} was used to generate 12 additional variations per system. This results in 60 generated clips. We also curated the top 15 high-quality outputs from the FramePack~\cite{zhang2025framepack} project page and supplemented the test set with 15 authentic videos, sourced from online platforms and our own recordings. This diverse test set allows us to evaluate the generalization of \textbf{DAVID-XR1} across various generation engines, modes, content styles, and real-world footage.

\paragraph{Evaluation Prompts}
For zero-shot evaluation of both SOTA commercial VLMMs and the Qwen2.5-VL-7B backbone (used by DAVID-XR1), we design prompts that include definitions for all labels used in our evidence annotations. Each model is required to output the complete \texttt{<think>…<evidence>…<answer>} structure: \texttt{<think>} provides step-by-step reasoning, \texttt{<evidence>} lists identified defect cues, and \texttt{<answer>} delivers the final verdict on whether the video is real or AI-generated. This setup evaluates the model’s ability to both detect AI-generated content and clearly articulate its reasoning. For the task-specific evaluation of DAVID-XR1, we use the same prompts employed during CoT-aware SFT (see Figure~\ref{fig:illustration}), as these consistently elicit the model’s full reasoning chain.

\paragraph{Evaluation Metrics}
To assess detection performance, we measure \emph{recall} for each category of out-of-domain synthetic and real videos, and report overall \emph{accuracy} to reflect comprehensive model performance. For explainable reasoning, we conduct manual evaluation using two metrics: (1) \emph{precision}, which measures whether each cue the model provides actually appears in the video and is a valid indicator of AI generation; and (2) \emph{diversity}, defined as the fraction of ground-truth evidence items (from all correct model responses) that the tested model retrieves. Diversity is a relative metric, with the denominator being the total number of correct evidence items identified by all evaluated models, thus reflecting the breadth of the model’s reasoning. All manual annotations are performed by our \textit{DAVID-X} labeler, as current models lack sufficient defect-understanding capabilities.

% For spatio-temporal localization, we uniformly assess whether the point the model marks on the specified frame truly falls within the described defect region: a full score indicates an exact hit, zero indicates a miss, and intermediate scores reflect partial alignment. Commercial baselines are judged entirely by human raters due to their inconsistent output formats; for all DAVID-XR1 variants, we instead use the VLMM to assist evaluation—overlaying a red dot at the \texttt{<point>} coordinate and asking the model to score it according to our criteria. By combining these multiple evaluation dimensions, we aim to faithfully reflect \textbf{DAVID-XR1}’s detection accuracy, explanation quality, and localization precision.

\begin{table}[t]
\centering
\caption{Evaluation results on out-of-distribution test dataset.}
\label{tab:main_results}
\resizebox{\textwidth}{!}{
\begin{tabular}{l|ccccccc|cc}
\toprule[2pt]
\multirow{3}{*}{\textbf{Model}} & \multicolumn{7}{c|}{\textbf{Detection}} & \multicolumn{2}{c}{\textbf{Explanation}} \\
\cline{2-8}\cline{9-10}
& \multicolumn{2}{c|}{\textbf{Next generation (↑)}} & \multicolumn{3}{c|}{\textbf{Out-of-Distribution (↑)}} & \multirow{2}{*}{\textbf{Real (↑)}} & \multirow{2}{*}{\textbf{Avg. (↑)}}&\multirow{2}{*}{\textbf{Precision (↑)}} & \multirow{2}{*}{{\textbf{Diversity (↑)}}} \\
\cline{2-6}
& Kling 2.0 & \multicolumn{1}{c|}{Pika v2.2} & Veo2 & Magi & \multicolumn{1}{c|}{FramePack} &  &  &  &  \\
\midrule[1pt]
GPT-4o                 & 20.0 & 53.3 & 20.0 & 33.3 & 33.3 & 73.3 & 38.9 & 15.33 & 14.40  \\
GPT‑4.1                & \underline{60.0}& 80.0 & \textbf{53.3} & 66.7 & \textbf{100} & \underline{80.0} & \underline{73.3} & 41.22 & \textbf{43.11} \\
Gemini-2.0-flash       & 13.3 & 60.0 & 20.0 & 40.0 & 86.7 & 60.0 & 46.7 & 22.67 & 11.73 \\
Gemini‑2.5‑pro         & 46.7 & \textbf{93.3} & 40.0 & \textbf{86.7} & \textbf{100} & 66.7 & 72.2 & \underline{49.33} & \underline{30.67} \\
\midrule[1pt]
Qwen2.5-VL-7B       & 6.7 & 26.7 & 13.3 & 0 & 26.7 & \textbf{86.7} & 26.7 & 3.20 & 3.87 \\
\rowcolor{green!15}
& \textbf{80.0} & \underline{86.7} & \underline{46.7} & \underline{80.0} & \underline{93.3} & 73.3 & \textbf{76.7} & \textbf{54.67} & 29.60 \\
\rowcolor{green!15}
\multirow{-2}{*}{\textbf{DAVID-XR1 (Ours)}}& \scriptsize\quad$+73.3$ & \scriptsize\quad$+60.0$ & \scriptsize\quad$+33.4$ & \scriptsize\quad$+80.0$ & \scriptsize\quad$+66.6$ & \scriptsize\quad$-13.3$ & \scriptsize\quad$+50.0$ & \scriptsize\quad$+51.47$ & \scriptsize\quad$+25.73$ \\
\bottomrule[2pt]
\end{tabular}
}
\end{table}

\subsection{Main Results}

Table \ref{tab:main_results} summarizes the performance of various VLMMs on the explainable AI-generated video detection task. Notably, our \textbf{DAVID-XR1} achieves a detection accuracy of 76.7\% and an explanation precision of 62.5\%, outperforming several SOTA general-purpose models. Closer inspection of the baselines reveals that recent models such as GPT-4.1 and the reasoning-focused Gemini-2.5-Pro generalize much better to our task than earlier architectures. Compared to its backbone, Qwen2.5-VL-7B, \textbf{DAVID-XR1}—after fine-tuning on \textit{DAVID-X}—demonstrates substantial improvements in both detection accuracy and explanatory capability. This confirms that our dataset effectively trains models not only to detect AI-generated content but also to provide justifications for their decisions. Furthermore, consistent performance across new generators, diverse model architectures, and different generation paradigms underscores the robustness of our approach and suggests that explainable reasoning enhances generalization.

It is also worth mentioning that GPT-4.1, thanks to its advanced visual reasoning abilities, achieves the highest diversity in evidence cues—surpassing all other models, including our \textbf{DAVID-XR1}. However, many of these cues are hallucinated or invalid, resulting in lower precision. The diversity score of \textbf{DAVID-XR1} is lower than that of GPT-4.1 and the teacher model Gemini-2.5-Pro, partly because we limited each sample to 1–3 defect cues and split examples with numerous cues into multiple entries during training, which slightly constrains output variety.

\subsection{Ablation and Case Studies}
\label{sec:ablation}

\begin{figure}[!h]
  \centering
  \includegraphics[width=1.0\linewidth]{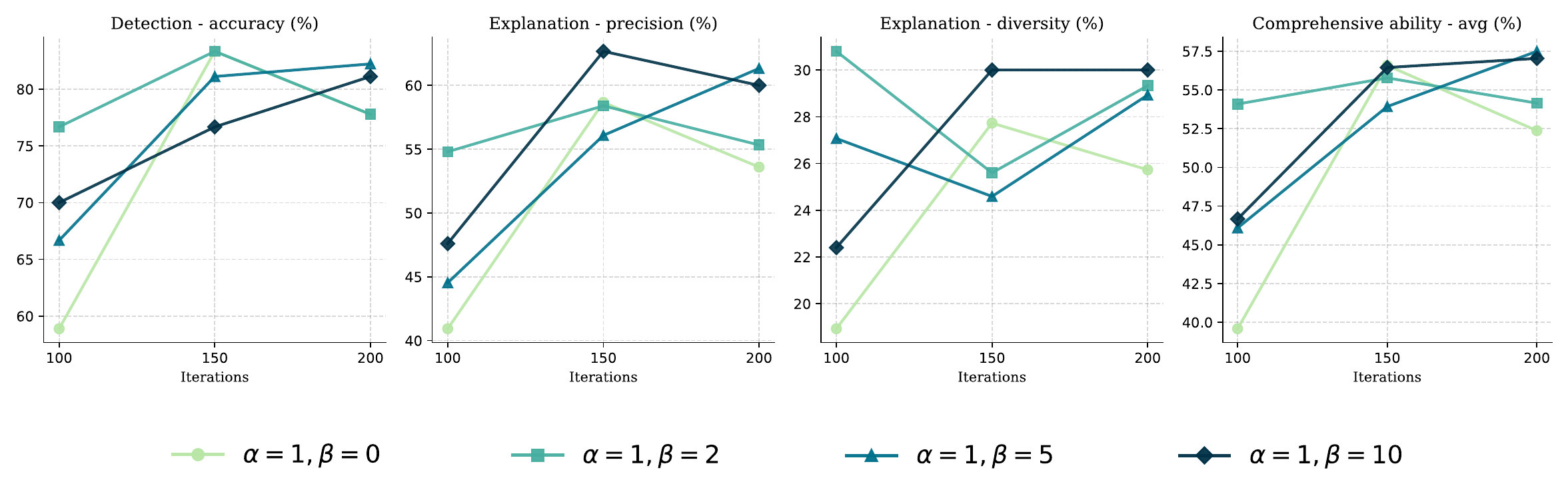}
  \caption{The detection and explaination performances of DAVId-XR1 under different hyperparameters ($\alpha$ and $\beta$) and fine-tuning iterations.}
  \vspace{-1.0em}
  \label{Fig:ablation}
\end{figure}

\paragraph{Ablation Study}
To assess how CoT-aware SFT performance evolves and how the loss-weight balance impacts detection and explanation, we conduct an ablation study across multiple training checkpoints and various $\alpha{/}\beta$ hyperparameter settings, where $\alpha$ and $\beta$ control the weights for language-model and classification objectives ($\beta = 0$ indicates pure language-model loss). Figure \ref{Fig:ablation} shows performance trends during fine-tuning.
The model rapidly improves in detection accuracy and interpretability during early epochs, with continued training reducing overfitting and smoothing the performance curve. Analysis of explanation-precision and explanation-diversity (the two central plots) shows that, across all loss-weight ratios, joint training consistently outperforms language-only training. Notably, an $\alpha:\beta$ ratio of $1:10$ yields substantial interpretability gains throughout training.
These results suggest that integrating chain-of-thought objectives into SFT accelerates and stabilizes learning while producing stronger, more reliable explanations. Further work is needed to identify the optimal balance between accuracy, interpretability, and generalization.

\begin{figure}[!t]
  \centering
  \includegraphics[width=0.98\linewidth]{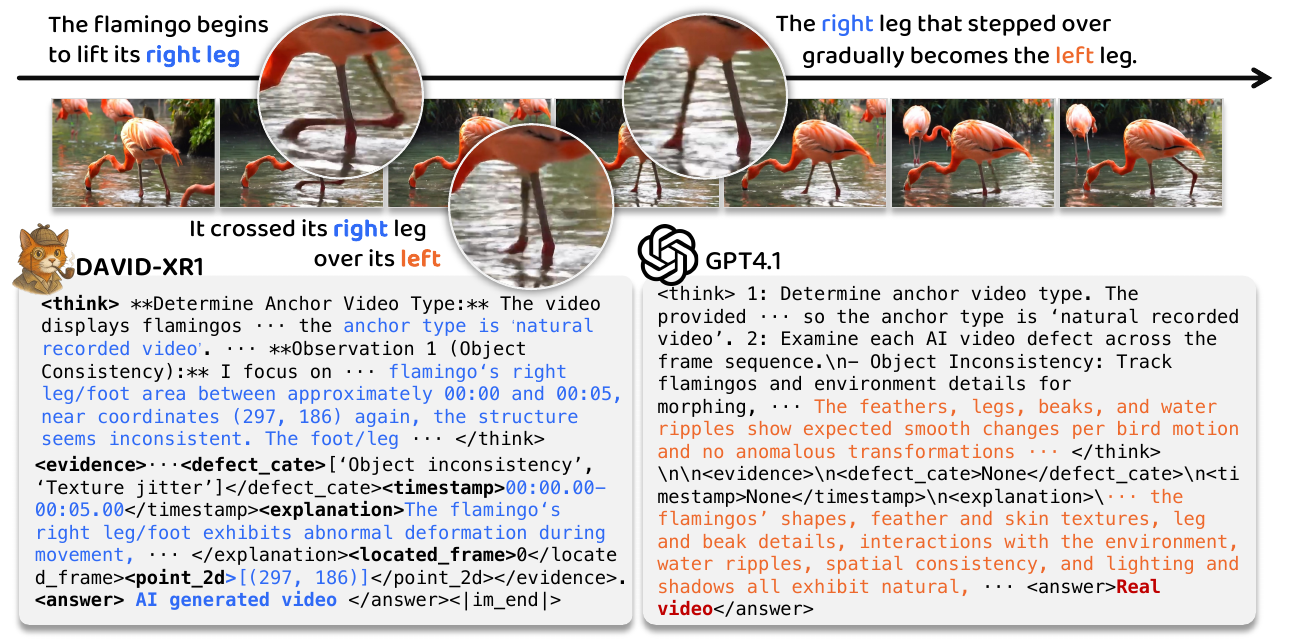}
  \caption{Example detection of an AI‐generated video (by Veo2) using our \textbf{DAVID-XR1} and GPT-4.1}
  \vspace{-1em}
  \label{Fig:Case_study}
\end{figure}

\paragraph{Case Study}
We select a representative clip generated by Veo2 to showcase our approach. As shown in Figure \ref{Fig:Case_study}, a flamingo’s right leg crosses over its left, briefly blurring and merging so that the right leg appears as the left. Although this sequence features multiple generative defects, their brief duration and small spatial extent make them extremely challenging to detect and localize.
When prompted to generate the full \texttt{<think>…<evidence>…<answer>} reasoning chain, both GPT-4.1 and \textbf{DAVID-XR1} produce coherent, explainable outputs, demonstrating that Chain-of-Thought distillation enables our model to match leading commercial models. Notably, \textbf{DAVID-XR1} accurately identifies the subtle defect in the flamingo’s legs, reflecting its mastery of domain-specific knowledge learned from \textit{DAVID-X}.

\section{Conclusion}

Driven by the need for explainable AI-generated video detection, we introduce the fine-grained annotation dataset \textbf{DAVID-X}, which captures defect categories, precise spatio-temporal localization, and detailed explanations. By distilling visual Chain-of-Thought (CoT) traces from multimodal reasoning models, we assemble structured reasoning data and use it to train an explainable reasoning model, \textbf{DAVID-XR1}, for AI-generated video detection. Extensive experiments demonstrate that DAVID-XR1 outperforms existing models in delivering evidence-based detection and explanations. Our results validate the feasibility of developing AI-generated video detectors that are not only accurate but also capable of providing detailed reasoning. In future work, we plan to explore reinforcement learning to further improve DAVID-XR1’s explanation precision, diversity, and detection accuracy.

\bibliographystyle{plainnat}
\bibliography{neurips_2025}

\newpage
\appendix

%%%%%%%%%%%%%%%%%%%%%%%%%%%%%%%%%%%%%%%%%%%%%%%%%%%%%%%%%%%%
\section{Keywords of Top-30 Prompt Clusters}
\label{app:impl}

\begin{figure}[!h]
  \centering
  \includegraphics[width=0.98\linewidth]{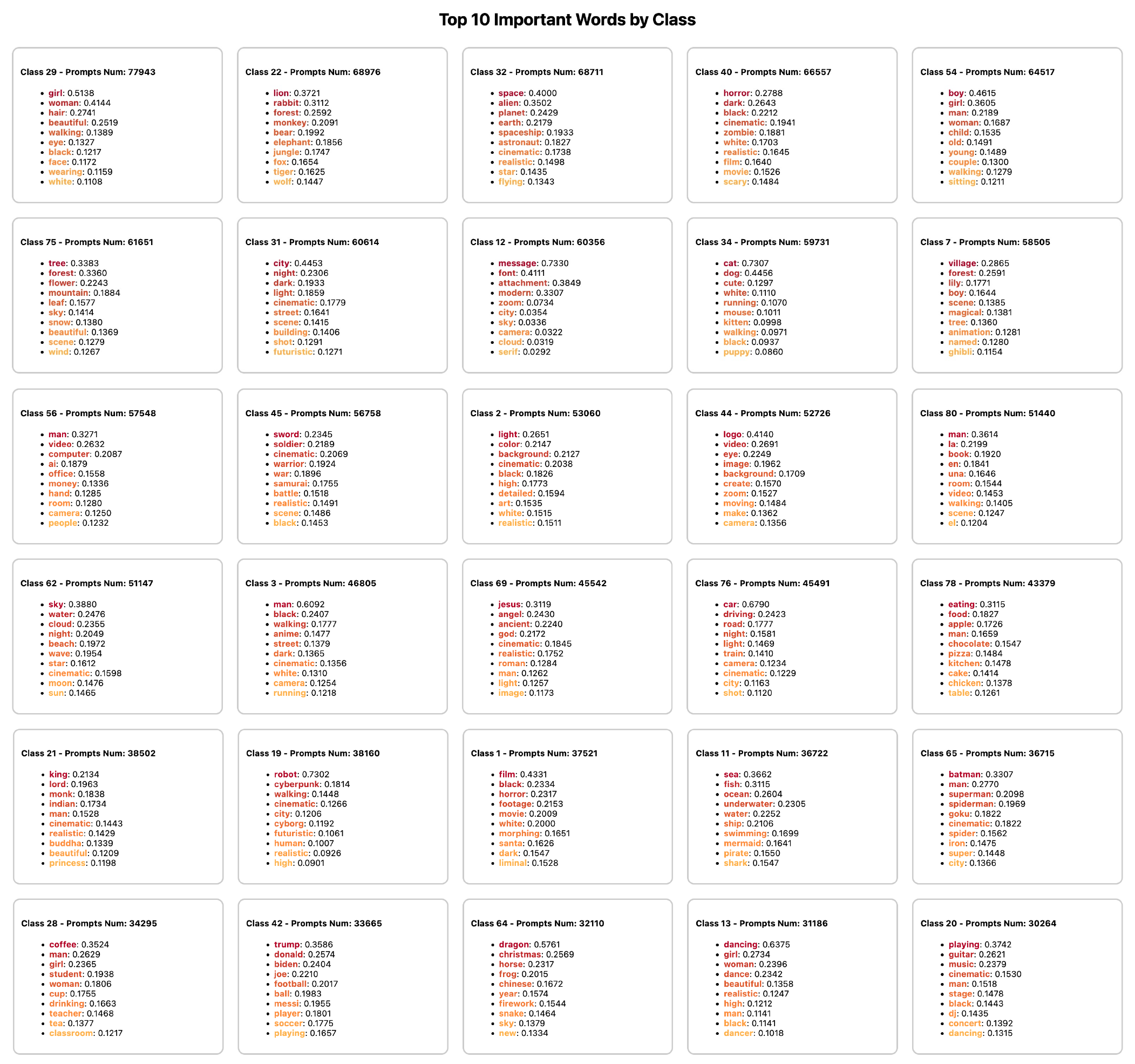}
  \caption{Keywords of top-30 prompt clusters}
  \label{Fig:Prompts_class}
\end{figure}

We cluster the real‐world generation prompts from VidProM \cite{wang2024vidprom} into 80 groups and retain the top 30, which together account for over 89 \% of all queries—thereby capturing the predominant user preferences. We then apply TF-IDF to each cluster to extract ten representative keywords (Figure \ref{Fig:Prompts_class}), which serve both as selection criteria for our sampled prompts and as reference anchors for LLM‐based prompt augmentation, ensuring our dataset mirrors prevailing generation trends.

\section{Zero-Shot Evaluation Prompts}

We include the zero-shot prompt used to evaluate both commercial VLMMs and Qwen2.5-VL-7B. The prompt embeds our full set of evidence-label definitions and simply asks for the \texttt{<think>···<evidence>···<answer>} chain, thereby testing both detection accuracy and reasoning clarity.

\begin{promptbox}
\texttt{Now, you are an AI-generated video identification assistant. Determine whether this video is AI-generated. First, you need to determine the anchor video type for this video: either 'natural recorded video' or 'handcrafted video'. 'natural recorded video' refers to footage captured directly by recording equipment with minimal post-processing; whereas 'handcrafted video' includes videos with significant post-production effects, or works from film, animation, and gaming. The tolerance level for identifying AI generation differs depending on which of these two types the video anchors to. Generally, there is a higher tolerance for potential artifacts when it is anchored to handcrafted video, since such videos do not need to conform to too many real-world physical laws. The following sections detail the intermediate results of the AI identification process, including which of the six types of AI generation defects were observed: \# Object Inconsistency: Refers to the failure of objects within an AI-generated video to maintain stable and coherent visual attributes (such as shape, size, color, or specific features) over time across frames. This often results in illogical transformations, distortion, loss/error of detail, or even complete degradation. \# Texture Jitter: Refers to the unnatural, subtle, high-frequency flickering, shifting, or distortion of surface texture details (especially in dense or repetitive areas) over time within AI-generated videos. This phenomenon is characterized by the instability of the texture pattern itself, rather than uniform noise, sometimes appearing as grid-like/mosaic disturbances, drifting of fine lines, or dynamic effects resembling "heat haze" or "crawling ants". (Distinct from the blocky artifacts or overall blurriness typically caused by compression or low resolution in conventional video). \# Interaction Anomaly: Refers to a defect in AI-generated videos where the distinct boundaries between two or more objects with similar visual attributes (e.g., shape, color) become blurred or confused when these objects visually overlap or are in close proximity within the camera's perspective, leading to an unnatural visual merging, sticking, or partial fusion of their forms. \# Movement Anomaly: Refers to the motion or action of objects or characters within an AI-generated video that appears unnatural, illogical, or physically/biologically implausible, like distorted trajectories and jerky or non-fluid movements. \# Space Anomaly: Refers to inconsistencies or illogical spatial relationships that emerge in AI-generated videos during dynamic camera movements, like disproportionate relative motion between foreground and background elements or visual discontinuities or misalignments where newly revealed scene areas fail to seamlessly connect with existing ones. \# Lighting Anomaly: Refers to the illogical, inconsistent, or physically implausible representation of light and shadow within an AI-generated video, like shadows whose direction, shape, or softness does not match apparent or implied light sources and unnatural shifts or flickering in light intensity or effects over time. The corresponding defect category is listed between '<defect\_cate>' and '</defect\_cate>'. Next is the time range where this defect occurs, marked between '<timestamp>' and '</timestamp>'; followed by a more specific spatial localization and description of the defect, where the textual explanation is enclosed between '<explanation>' and '</explanation>'. The defect is then annotated with a 2‑D point ('point\_2d'): first specify the frame that contains this point by placing its frame number between '<located\_frame>' and '</located\_frame>'—this identifies the exact frame on which the point should be drawn—then give all point’s (x, y) coordinates between '<point\_2d>' and '</point\_2d>'. Please make sure that the point coordinates inside <point\_2d></point\_2d> should be normalized to the range 0–1000. If after your analysis you believe this is a real video with no AI generated defects, you do not need to give the spatiotemporal location of the defects. Only fill in the explanation between <explanation> and </explanation> to explain why you believe this is a real video, and use 'None' as a placeholder for the other sections. The final answer should be either 'AI generated video' or 'Real video', placed between '<answer>' and '</answer>'. You need to think carefully and provide more standardized and readable fine-grained identification evidence, and eventually provide your conclusion. 1) Output all your chain of thought (CoT process) inside <think></think>; 2) then, inside <evidence></evidence>, give the above evidence points for each defect in turn, including <defect\_cate></defect\_cate>, <timestamp></timestamp>, <explanation></explanation>, <located\_frame></located\_frame>, and <point\_2d></point\_2d>; 3) finally, provide your identification result within <answer></answer>.}
\end{promptbox}

\section{DAVID-X Annotation Tool}

We developed a custom annotation tool for AI-generated video detection using Gradio\cite{abid2019gradio} and Sam2\cite{ravi2024sam}. All spatial points were manually refined through multiple tries, including both positive and negative samples to achieve precise defect region masks. We look forward to employing vision–language models with enhanced spatio-temporal segmentation capabilities in future work to achieve explainable reasoning with even end2end finer-grained localization.

\begin{figure}[!h]
  \centering
  \includegraphics[width=0.98\linewidth]{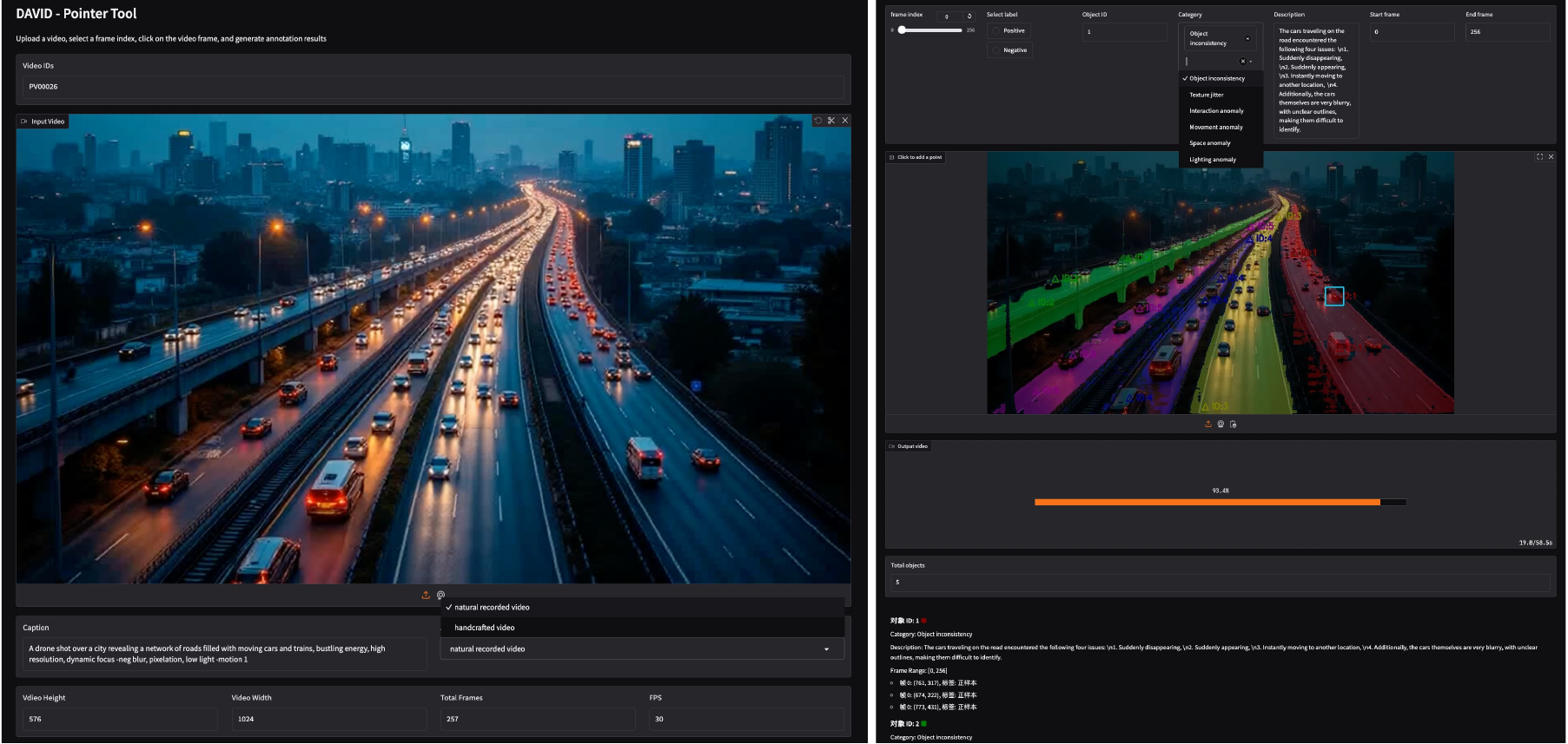}
  \caption{DAVID-X Annotation Tool}
  \label{Fig:Annotator}
\end{figure}

\section{DAVID-XR1 Demo}

We present a demonstration of \textbf{DAVID-XR1} in real-world deployment with Chainlit and include the demo video in the supplementary materials.

\begin{figure}[!h]
  \centering
  \includegraphics[width=0.98\linewidth]{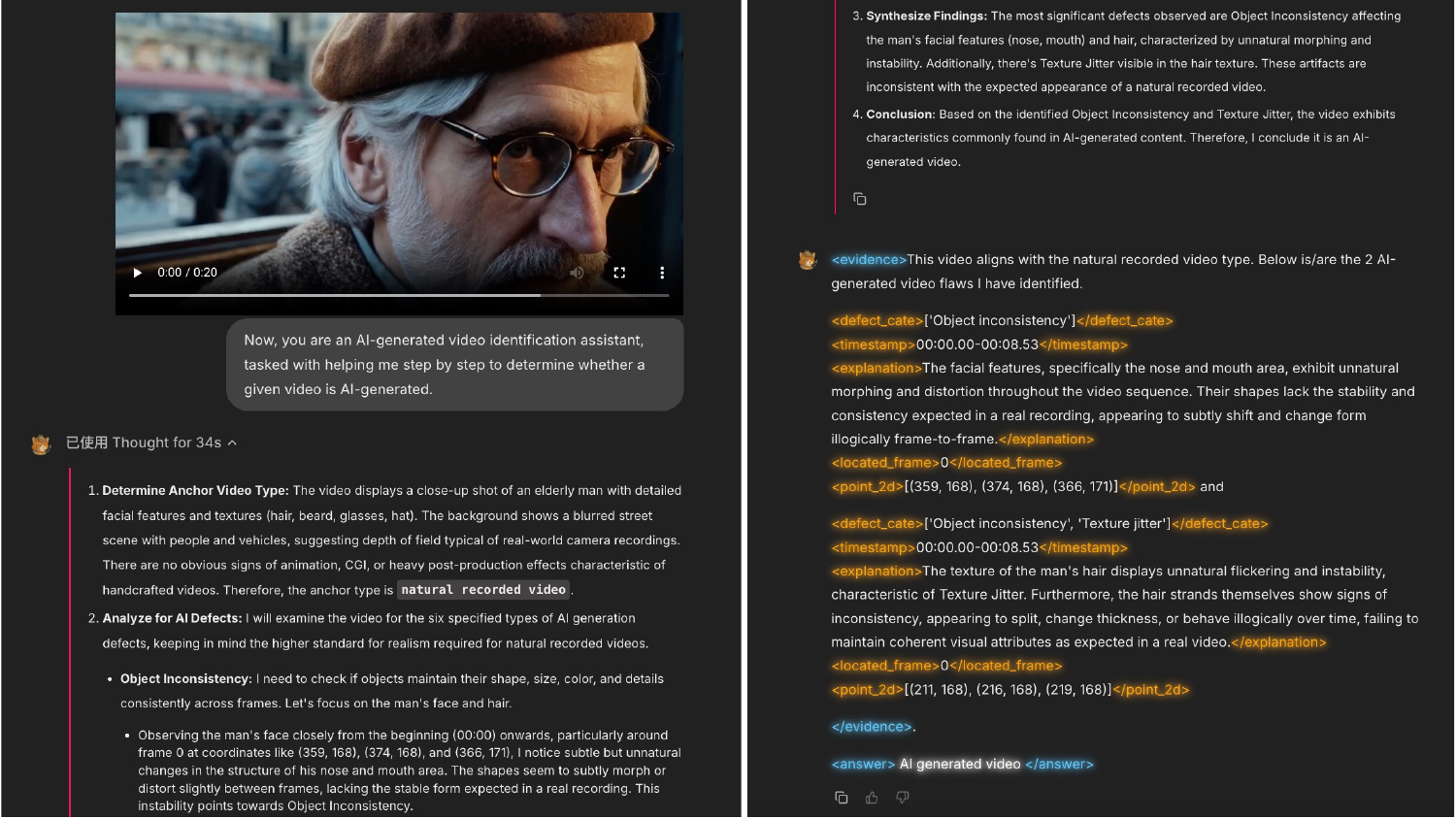}
  \caption{DAVID-XR1 Demo}
  \label{Fig:Demo}
\end{figure}

%%%%%%%%%%%%%%%%%%%%%%%%%%%%%%%%%%%%%%%%%%%%%%%%%%%%%%%%%%%%

\FloatBarrier 

%%%%%%%%%%%%%%%%%%%%%%%%%%%%%%%%%%%%%%%%%%%%%%%%%%%%%%%%%%%%

\end{document}